\documentclass[10pt,twocolumn,letterpaper]{article}

\usepackage[numbers]{natbib}
\usepackage{cvpr}
\usepackage{times}
\usepackage{epsfig}
\usepackage{graphicx}
\usepackage{amsmath}
\usepackage{amssymb}
\usepackage{multirow}
\usepackage{bm}
\usepackage{booktabs}
\usepackage{subfig}
\usepackage[colorlinks,linkcolor=blue]{hyperref}

\cvprfinalcopy % *** Uncomment this line for the final submission

\def\cvprPaperID{****} % *** Enter the CVPR Paper ID here

\ifcvprfinal\fi
\begin{document}

%%%%%%%%% TITLE
%%%%%%%%% TITLE
\title{Fast Object Placement Assessment}

\author{$\textnormal{Li Niu}^{*}$, $\textnormal{Qingyang Liu}^{\dag}$, $\textnormal{Zhenchen Liu}^{*}$, $\textnormal{Jiangtong Li}^{*}$\\
$^*$ Shanghai Jiao Tong University\,\,
$\dag$ Beijing Institute of Technology \\
}

\maketitle
\thispagestyle{empty}

%%%%%%%%% ABSTRACT
\begin{abstract}
Object placement assessment (OPA) aims to predict the rationality score of a composite image in terms of the placement (\emph{e.g.}, scale, location) of inserted foreground object. However, given a pair of scaled foreground and background, to enumerate all the reasonable locations, existing OPA model needs to place the foreground at each location on the background and pass the obtained composite image through the model one at a time, which is very time-consuming. In this work, we investigate a new task named as fast OPA. Specifically, provided with a scaled foreground and a background, we only pass them through the model once and predict the rationality scores for all locations. To accomplish this task, we propose a pioneering fast OPA model with several innovations (\emph{i.e.}, foreground dynamic filter, background prior transfer, and composite feature mimicking) to bridge the performance gap between slow OPA model and fast OPA model. Extensive experiments on OPA dataset show that our proposed fast OPA model performs on par with slow OPA model but runs significantly faster. 
\end{abstract}

%%%%%%%%% BODY TEXT

\section{Introduction} \label{sec:intro}

Image composition aims to cut the foreground from one image and paste it on another image to form a realistic-looking composite image. As a common image editing operation, image composition has a pivotal role in augmented reality, artistic creation, and automatic advertising \cite{WhereandWhoTan2018,MISCWeng2020,WhatWhereZhang2020}. The main challenges to create a realistic-looking composite can be summarized as the inconsistency between foreground and background~\cite{TowardRealisticChen2019,niu2021making}, which might be caused by the improper color/illumination, scale, location, perspective of foreground or the missing impact of foreground on the background (\emph{e.g.}, shadow, reflection). Most previous works on image composition targeted at solving one or several issues, like image harmonization \cite{TsaiDIHarmonization2017,CongDoveNet2020,IHCISSAMCun2020}, shadow generation \cite{ARShadowGANLiu2020,hong2021shadow}, and object placement learning \cite{SyntheticTripathi2019,LearningObjPlaZhang2020,WhereandWhoTan2018}.

\begin{figure*}[ht]
\centering
\includegraphics[width=0.99\textwidth]{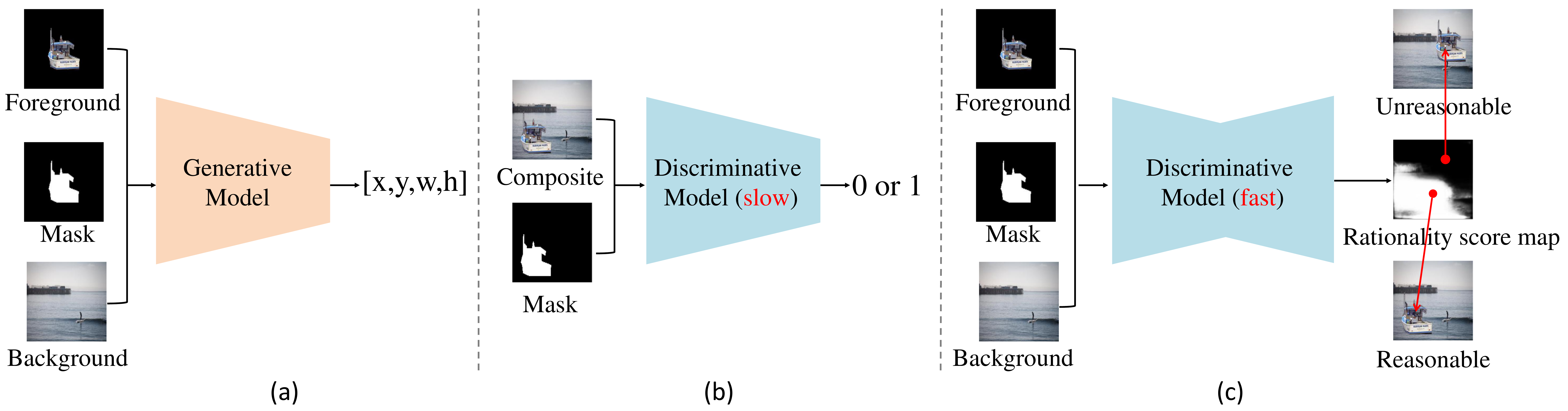} 
.
\caption{The comparison among generative model (a), slow object placement assessment model (b), and fast object placement assessment model (c). In (a), given the foreground, foreground mask, and background, the model generates a reasonable placement (\emph{e.g.}, location (x,y) and scale (w,h)) for the foreground. In (b), given the composite image and composite foreground mask, the model predicts a rationality score. In (c), with the same input as (a), the model predicts a rationality score map containing the rationality scores for all locations.  }
\label{fig:comparison}
\end{figure*}

In this work, we focus on object placement learning, which targets at finding reasonable location and scale of the foreground object. Object placement has a wide range of application scenarios. For example, in artistic creation, this technique could provide designers with feedback and make recommendation for them when they are placing objects. Another application is automatic advertising, which aims to help advertisers with automatic product insertion in the background scene~\cite{WhatWhereZhang2020}.
Object placement learning can be treated as either a generative task or a discriminative task. The generative task aims to find one or multiple reasonable placements for the foreground object, as shown in Figure~\ref{fig:comparison}(a).  Previous methods \cite{chen2021geosim,STDODISGeorgakis2017,LSCPRemez2018,LearningObjPlaZhang2020,SynthesizingGeorgakis2017} take in a foreground and a background, predicting one or multiple plausible placements for the foreground object. 
However, they cannot enumerate all the reasonable placements efficiently and effectively. Alternatively, the discriminative task, also named as object placement assessment (OPA) \cite{liu2021opa}, aims to verify whether the foreground object placement in a composite image is plausible. 
Recently, \cite{liu2021opa} released an Object Placement Assessment (OPA) dataset containing abundant composite images with annotated rationality labels, which paves the way for OPA task. 
\cite{liu2021opa} also proposed a simple baseline method simOPA, which predicts the rationality score for a composite image as shown in Figure~\ref{fig:comparison}(b).

One severe limitation of simOPA is inefficiency. Specifically, to enumerate all the reasonable placements, it needs to place the foreground at each location on the background and pass the obtained composite image through the model one at a time, which is very time-consuming. In this paper, we ask the following question: \emph{given a scaled foreground and a background image, can we go through a model only once and predict the rationality scores for all locations?} This new task is depicted in Figure~\ref{fig:comparison}(c). For ease of description, we refer to simOPA as Slow OPA (SOPA) model and the dense prediction model in Figure~\ref{fig:comparison}(c) as fast OPA (FOPA) model. 

Since there is no previous literature on FOPA, we start constructing our FOPA model from scratch. One intuitive way is extracting the pixel-wise feature of background and the feature of foreground, based on which pixel-wise rationality score can be predicted by integrating the foreground feature and pixel-wise background feature. 
Specifically, we use U-Net structure \cite{ronneberger2015u} to extract the background feature map and an encoder to extract the foreground feature. Inspired by recent works \cite{jia2016dynamic,Sharma_2018_CVPR,pang2020hierarchical} using dynamic filter, we use foreground feature to generate dynamic filters, which are applied to the background feature map to produce the output feature map. Finally, we make pixel-wise prediction based on the output feature map and generate a rationality score map. 

To boost the performance of FOPA model, we propose two techniques to transfer knowledge from SOPA model to FOPA model.
The first one is background prior transfer. Since SOPA model captures rich prior knowledge to assess the rationality of object placement, we replace the background encoder parameters with SOPA model parameters and fix them during training. In this way, the background encoder can leverage the background prior knowledge encapsulated in the SOPA model. 
The second one is composite feature mimicking. Precisely, we enforce each pixel-wise feature in the output feature map to be close to the  image-level feature of the corresponding composite image (place the foreground at this location on the background) extracted by SOPA model. In this way, the knowledge of composite image features, especially the subtle difference caused by placing the foreground at various locations, can be transferred from SOPA model to FOPA model. 

Given a pair of scaled foreground and background, our model can predict the rationality scores for all locations and obtain a large number of realistic composite images efficiently.
\emph{To further accelerate processing different foreground scales, we also explore one-hot scale encoding (see Section~\ref{sec:network_architecture}) to avoid passing through the foreground encoder for multiple times. }
We evaluate our proposed FOPA model on OPA dataset  \cite{liu2021opa} and target at $256\times 256$ rationality score map, which has the same size as the input background. We observe that our FOPA model achieves comparable results with SOPA model for object placement assessment, but consumes much less resource than SOPA model for producing a complete rationality score map.  
Our contributions can be summarized as follows: 
\begin{itemize}
\item  We define a new task named fast object placement assessment (FOPA) and design a novel network structure for this task. Given a scaled foreground and a background image, we can go through the model only once and predict the rationality scores for all locations. 
\item We propose several innovations to bridge the gap between slow OPA and fast OPA, including foreground dynamic filter, background prior transfer, and composite feature mimicking. 
\item Extensive experiments on OPA dataset show that our proposed FOPA model performs on par with SOPA model but runs significantly faster.
\end{itemize}

\section{Related Work}

\subsection{Image Composition}
The goal of image composition is combining foreground and background as a realistic composite image. However, the fidelity and quality of composite images may be severely harmed by the inconsistency between foreground and background \cite{TowardRealisticChen2019,niu2021making}. For example, the foreground and background have incompatible color and illumination statistics. To tackle this issue, image harmonization \cite{TsaiDIHarmonization2017,CongDoveNet2020,Ling2021RegionawareAI,guo2021intrinsic,IHCISSAMCun2020,sofiiuk2021foreground} aims to adjust the color and illumination of foreground to make it compatible with the background, yielding a harmonious composite image. Besides, the inserted foreground may also have impact on the background such as casting shadow or reflection, so some works \cite{ARShadowGANLiu2020,hong2021shadow} attempted to generate plausible shadow for the inserted foreground.
Another issue is that the location and scale of the inserted foreground might be unreasonable. 
To tackle this issue, object placement methods~\cite{MVCCPDvornik2018,WhereandWhoTan2018,WhatWhereZhang2020,ContextawareLee2018,SyntheticTripathi2019,LearningObjPlaZhang2020} attempted to seek for the plausible location and scale for the inserted foreground, which will be detailed next. 

\subsection{Object Placement Learning}
\label{subsection:Object_Placement_Learning}
Existing works on object placement learning can be roughly divided into generative model and discriminative model. The generative model aims to generate reasonable placements for the foreground to be pasted on the background. For example, the methods \cite{STDODISGeorgakis2017,LSCPRemez2018,InstaBoostFang2019,SyntheticTripathi2019} predicted one plausible placement (\emph{e.g.}, bounding box, warping parameters) for the foreground based on a pair of foreground and background. \cite{LearningObjPlaZhang2020} predicted multiple plausible placements by injecting a random vector into the generator. Nevertheless, they are unable to predict all the reasonable placements efficiently and effectively. Different from generative model, the discriminative model aims to assess whether a placement is plausible, which is referred to as object placement assessment \cite{liu2021opa}. The discriminative model proposed in \cite{liu2021opa} is simply a binary classification model which can only evaluate the rationality of one composite image corresponding to one object placement at a time. Thus, it will take heavy cost to find all the reasonable placements. Therefore, given a pair of scaled foreground and background, we design a fast object placement assessment model to obtain all reasonable locations with just a single pass through the network.

\subsection{Dynamic Filter}
The concept of dynamic filter \cite{jia2016dynamic, Sharma_2018_CVPR, pang2020hierarchical} was proposed to learn a variety of filtering neighborhood and filtering operations. 
One type of approaches dynamically adjust the filtering neighborhoods by adaptive dilation factors \cite{zhang2017scale} or estimating the receptive fields \cite{tabernik2020spatially}.
Another type of approaches first use a convolutional network or a multilayer perceptron to produce filters conditioned on an input, and then apply the generated filters to another input \cite{yang2019condconv,wang2019carafe,su2019pixel,tang2020learning,ding2020learning,chen2020dynamic,wang2020advances,tian2020conditional,zhou2021decoupled}. In contrast with the traditional convolutional layer which fixes the learned filters after training, the dynamic filter network generates filters dynamically conditioned on an input to enhance the representation in a self-learning manner.
Our method belongs to the second type of approaches, which have been used in image classification \cite{yang2019condconv, chen2020dynamic, zhou2021decoupled}, object detection \cite{wang2019carafe, ding2020learning}, instance segmentation \cite{wang2019carafe, wang2020advances, tian2020conditional}, semantic segmentation \cite{wang2019carafe,su2019pixel}, image inpainting \cite{wang2019carafe}, and depth completion \cite{tang2020learning, zhou2021decoupled}.
In this work, we generate convolution filters dynamically based on the foreground and apply the filters to each location of background feature map, which has never been explored before.

\subsection{Feature Distillation}
Knowledge distillation \cite{hinton2015distilling} targets at transferring knowledge from a stronger teacher network to a weaker student network, where teacher network is usually a large pretrained network to provide smoother supervision towards the student network. Based on the type of knowledge, the knowledge distillation can be divided into three groups \cite{gou2021knowledge}: response-based \cite{hinton2015distilling}, relation-based \cite{yim2017gift}, and feature-based \cite{Li_2017_CVPR} .
Response-based knowledge distillation usually uses the neural response of the last output layer of the teacher model to supervise the student model \cite{hinton2015distilling, zhang2019fast}.
Relation-based knowledge distillation \cite{yim2017gift, passalis2020heterogeneous} pays more attention to relationships between different model layers or data samples.
Feature-based knowledge distillation \cite{RomeroBKCGB14, wang2020exclusivity,wang2021distilling} targets at both the output of the last layer and the output of intermediate layers to supervise the training of the student model, where the main idea is to directly match the feature activations of the teacher and the student.
Our method belongs to the feature-based knowledge distillation, which has been used in image classification \cite{RomeroBKCGB14, kim2018paraphrasing, heo2019knowledge}, semantic segmentation \cite{liu2019structured, shu2021channel}, object detection \cite{chen2017learning, chawla2021data}, and face recognition \cite{wang2020exclusivity}.
In this work, we employ feature distillation/mimicking to help our FOPA model learn from pretrained SOPA model to bridge their performance gap.

\section{Our Method}

\subsection{Network Architecture} \label{sec:network_architecture}
\begin{figure*}[t]
\centering
\includegraphics[width=0.95\textwidth]{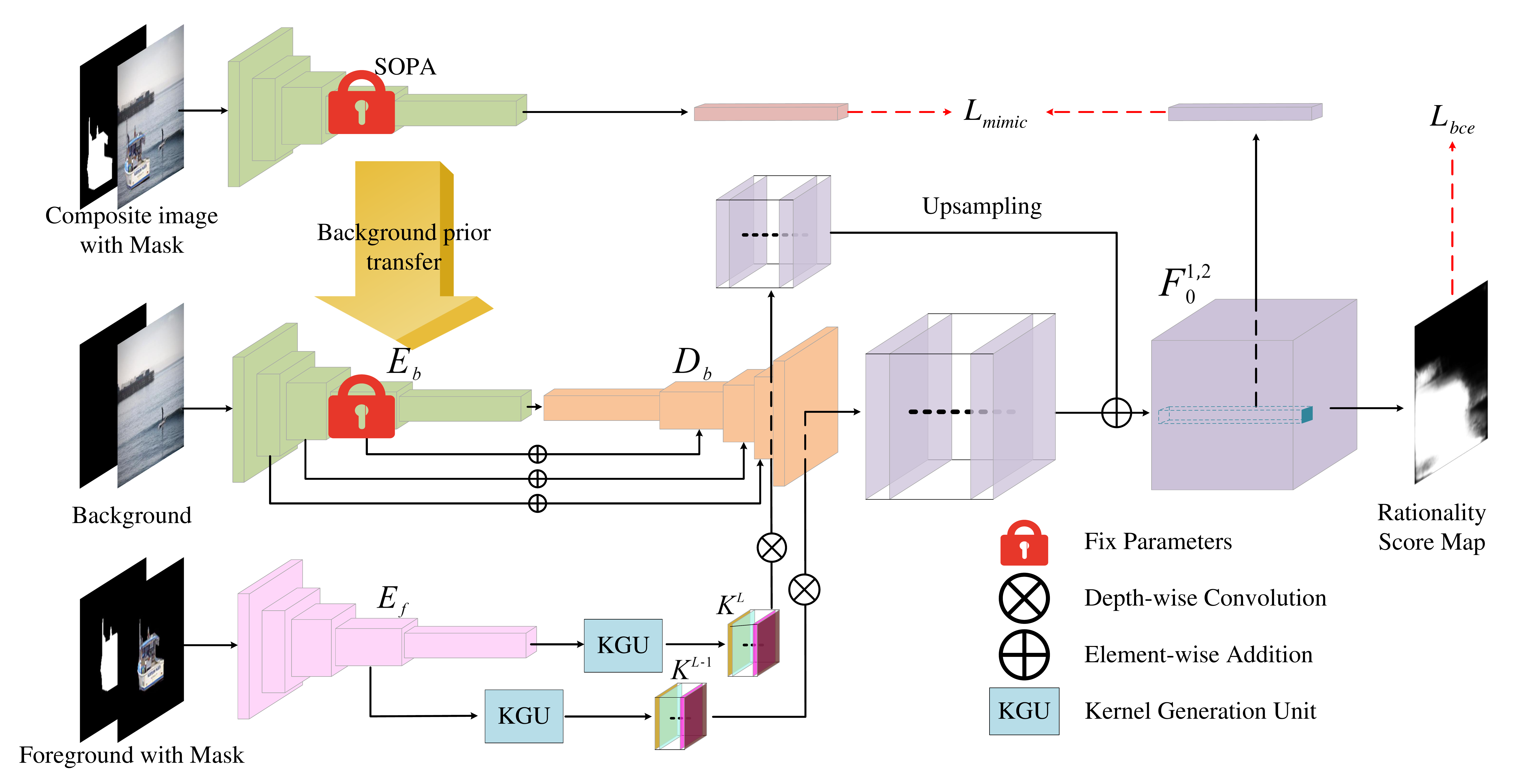}
\caption{Network architecture of our fast object placement assessment (FOPA) model. We first train a slow object placement assessment (SOPA) model and use it as the background encoder $E_b$. Our FOPA model consists of foreground branch ($E_f$) and background branch ($E_b$, $D_b$). We use foreground feature to generate dynamic filters, which act upon background feature map to predict rationality score map.}
\label{fig:Network_architecture}
\end{figure*}

An overview of our network is shown in Figure \ref{fig:Network_architecture}. Given a foreground object $\mathbf{O}$, the corresponding foreground mask $\mathbf{M}$, and a background image $\mathbf{B}$, we expect to predict rationality score map $\mathbf{R}$, in which each entry $r_{i,j}$ means the rationality score of the composite image obtained by placing the center of foreground object $\mathbf{O}$ at the location $(i,j)$ on the background $\mathbf{B}$. 
Since our task needs to produce pixel-wise classification results, which is analogous to segmentation task, one natural thought is using popular encoder-decoder structure (\emph{e.g.}, U-Net \cite{ronneberger2015u}) to make dense predictions. 

In detail, given a pair of foreground object of expected scale and background, we place the scaled foreground object $\mathbf{O}$ and its mask $\mathbf{M}$ at the centre of a black image with the same size as background $\mathbf{B}$, yielding  $\mathbf{O}'$ and $\mathbf{M}'$. \emph{In this way, the scale information of foreground is encoded in the input. To accelerate processing different scales, we also develop an accelerated version, in which we extract the feature of full-scale foreground and concatenate it with one-hot scale encoding.} The details of accelerated version will be introduced in Supplementary. 
Then, we resize $\mathbf{O}'$, $\mathbf{M}'$, and $\mathbf{B}$ to the network input size $H\times W$ (\emph{e.g.}, $256 \times 256$). Our model has a foreground branch and a background branch. The background branch is U-Net structure \cite{ronneberger2015u} with an encoder $E_{b}$ and a decoder $D_{b}$ to extract background feature map from background image $\mathbf{B}$. The foreground branch is an encoder $E_{f}$ to extract foreground feature from the concatenated foreground image $\mathbf{O}'$ and its mask $\mathbf{M}'$. We adopt ResNet18 \cite{he2016deep} for $\{E_{f}, E_{b}\}$ and the decoder structure in \cite{ronneberger2015u} for $D_{b}$. %, which gradually recovers the feature map back to the input resolution with skip connections from the background encoder $E_{b}$. 

%Through foreground branch and background branch, we can obtain the foreground feature and background feature map. 
To simulate the process of pasting the foreground at each location on the background, we tend to integrate the foreground feature with pixel-wise background feature to predict a rationality score for each location. Inspired by recent works \cite{Sharma_2018_CVPR,pang2020hierarchical}, we adopt dynamic filter as the integration scheme.
Specifically, we use foreground feature to generate dynamic filters, which act upon each location of background feature map.

\subsection{Foreground Dynamic Filter}
As shown in Figure \ref{fig:Network_architecture}, our dynamic filter module consists of a Kernel Generation Unit (KGU) which produces kernels conditioned on the foreground feature, and a dynamic filtering layer which applies the generated kernels to the background feature map. The structure of KGU is adopted from \cite{jia2016dynamic}. 
For efficiency, we generate depth-wise convolutional filters \cite{guo2019depthwise}.

Through the foreground encoder $E_f$, we obtain the intermediate features $\{\mathbf{f}_f^l|_{l=1}^L\}$ ($L=4$) from high resolution to low resolution. Through the background decoder $D_{b}$, we obtain the intermediate features $\{\mathbf{F}_b^l|_{l=1}^L\}$ from high resolution to low resolution. As the first attempt, we apply the dynamic filters generated from the deepest foreground feature $\mathbf{f}^{L}_f$  to the last background feature map $\mathbf{F}^{1}_b$. Specifically, we feed $\mathbf{f}^{L}_f$ into KGU to generate depth-wise convolutional filters $\mathbf{K}^{L} \in \mathcal{R}^{k\times k\times d}$, in which $k=3$ is the kernel size and $d$ is the channel dimension.
Then, we apply  $\mathbf{K}^{L}$ to $\mathbf{F}^{1}_b \in \mathcal{R}^{H\times W\times d}$ with each $k\times k$ convolution filter acting upon each channel in $\mathbf{F}^{1}_b$, producing the output feature map $\mathbf{F}^1_{o}$. 

To further leverage the features of different scales, we extend the above single-scale dynamic filter to multi-scale dynamic filter. By taking two-scale dynamic filter as an example, we use $\mathbf{f}^{L-1}_f$ and $\mathbf{f}^{L}_f$ to generate  $\mathbf{K}^{L-1}$ and $\mathbf{K}^{L}$ respectively, which are applied to $\mathbf{F}^{1}_b \in \mathcal{R}^{H\times W\times d}$ and $\mathbf{F}^{2}_b \in \mathcal{R}^{\frac{H}{2}\times \frac{W}{2}\times d}$ respectively. 
%For the targets of multi-scale dynamic filters, we have also tried an alternative way (applying $\mathbf{K}^{L-1}$ and $\mathbf{K}^{L}$ to  $\mathbf{F}^{1}_b$ and $\mathbf{F}^{2}_b$ respectively), but the performance is slightly worse (see Section XXX).
After obtaining two output feature maps $\mathbf{F}^{1}_{o}\in \mathcal{R}^{H\times W\times d}$ and $\mathbf{F}^{2}_{o}\in \mathcal{R}^{\frac{H}{2}\times \frac{W}{2}\times d}$, we add up $\mathbf{F}^{1}_{o}$ and upsampled $\mathbf{F}^{2}_{o}$ to construct the new feature map $\mathbf{F}^{1,2}_{o}$. 

\subsection{Background Prior Transfer}
%Intuitively, it might be helpful to transfer information from previously learned task to a different yet related task \cite{liu2021opa}.
Recall that SOPA model takes in a composite image and its composite foreground mask to predict a rationality score.
We conjecture that SOPA model captures rich prior knowledge (\emph{e.g.}, semantic information, geometric information) to assess the rationality of object placement. Hence, we consider replacing the background encoder parameters of $E_b$ with SOPA model parameters.
The only difference is that the input of $E_b$ is pure background without composite foreground, so it is meaningless to concatenate composite foreground mask. Hence, we replace the composite foreground mask with an all-zero mask, which indicates the absence of composite foreground. 

Next, we provide some more in-depth analyses for background prior transfer. On the one hand, we conjecture that SOPA model can extract rich information from composite image which is useful for assessing the rationality of object placement. For example, as shown in the rationality score  map  of  SOPA  (see Figure~\ref{fig:Qualitative_Analyses} and Supplementary), SOPA  can  learn both semantic information (\emph{e.g.}, suitable semantic region for the foreground)  and  geometric  information  (\emph{e.g.}, flat surface). On the other hand, we conjecture that SOPA model can capture the relation between foreground and background based on the composite foreground mask. In our background encoder, since the input mask is an empty mask, the function of background encoder $E_b$ should be from the first aspect, that is, extracting rich background information (\emph{e.g.}, semantic information and geometric information) which is useful for  object placement assessment.

During training, we adopt a two-stage training strategy. First, we train a SOPA model. Then, we initialize $E_b$ with the trained SOPA model and freeze $E_b$ when training our FOPA model. We have also tried updating $E_b$ in the second training stage (see Section~\ref{section:Ablation_Studies}), but we witness significant performance drop probably due to the overfitting issue.

\subsection{Composite Feature Mimicking}
Inspired by previous literature on feature mimicking \cite{Li_2017_CVPR,wang2021distilling}, we incorporate a feature mimicking loss between the FOPA model and the SOPA model. Previous works \cite{Li_2017_CVPR,wang2021distilling} simply performed mimicking between the features of the same image/region extracted from different models. Here, we perform feature mimicking in an innovative manner. Specifically, we use each pixel-wise output feature from our FOPA model to mimic the image-level feature of the corresponding composite image from SOPA model, because each pixel on the output feature map corresponds to a composite image obtained by pasting the foreground at this location.  

One practical problem is that given a scaled foreground and a background, our used OPA dataset \cite{liu2021opa} only provides sparse annotations for a small proportion of pixels, so we only consider the set of annotated pixels $\mathcal{S}$. When training our FOPA model, we can obtain the pixel-wise output features for the annotated pixels $(x,y)\in \mathcal{S}$. We denote the pixel-wise feature at pixel $(x,y)$ in $\mathbf{F}_{o}^{1,2}$ as $\mathbf{f}^{1,2}_{o}(x,y) \in \mathcal{R}^{d}$. In the meanwhile, we also use trained SOPA model to extract the image-level feature $\hat{\mathbf{f}}_{x,y}\in \mathcal{R}^{\hat{d}}$ from the corresponding composite image. Since $d$ and $\hat{d}$ may be different,  we apply a fully-connected layer to $\mathbf{f}^{1,2}_{o}(x,y)$ to obtain $\hat{\mathbf{f}}^{1,2}_{o}(x,y) \in \mathcal{R}^{\hat{d}}$ to match the feature dimension. The feature mimicking loss is enforced between $\hat{\mathbf{f}}^{1,2}_{o}(x,y)$ and  $\hat{\mathbf{f}}_{x,y}$ as follows,
\begin{eqnarray}
L_{mimic}=\sum_{(x,y)\in \mathcal{S}}\|\hat{\mathbf{f}}^{1,2}_{o}(x,y)-\hat{\mathbf{f}}_{x,y}\|^2.
\end{eqnarray}

\emph{Note that our key point is transferring from slow model to fast model, rather than transferring from large model to small model.} 
The input of slow model SOPA is concrete composite image, so SOPA feature can capture the detailed and accurate information of composite image. In contrast, the input of fast model FOPA is foreground and background, so FOPA needs to imagine the feature of each composite image by integrating the foreground feature and each pixel-level background feature, which is very challenging. 
Therefore, we assume that the detailed and accurate information in SOPA feature can provide useful guidance for FOPA. 
%we need to pass the composite image through the SOPA model one at a time to acquire composite image features, which are actually more expressive than the pixel-wise output features from FOPA model.
Another interpretation is that the background encoder $E_b$ extracts the background feature, while the decoder $D_b$ together with foreground dynamic filter learns the feature variation caused by pasting the foreground on the background to simulate composite image features.  

Apart from the feature mimicking loss, the pixel-wise output feature $\mathbf{f}^{1,2}_{o}(x,y)$ is also supervised by the ground-truth rationality label (0 for unreasonable and 1 for reasonable)
with a binary cross-entropy loss:
\begin{eqnarray}
L_{bce}=-\sum_{(x,y)\in \mathcal{S}} log(p_c(\mathbf{f}^{1,2}_{o}(x,y))),
\end{eqnarray}
in which $p_c(\cdot)$ is the prediction score for the ground-truth label $c$. For the predicted rationality score map $\mathbf{R}$, each entry is $r_{x,y}=p_1(\mathbf{f}^{1,2}_{o}(x,y))$.
In summary, the total loss function of our FOPA model is
\begin{eqnarray} \label{eqn:total_loss}
L_{total}=L_{bce}+\lambda L_{mimic},
\end{eqnarray}
in which the trade-off parameter $\lambda$ is set as 16 via cross-validation.

\section{Experiments}
In this section, we investigate the effectiveness of each component and compare our FOPA model with SOPA model. We also show that our method can help generate realistic composite image. 

\setlength{\tabcolsep}{1.7mm}
\begin{table*}[t]
\caption{Ablation studies of our FOPA  method. In \emph{dynamic filter}, $\dagger$ means concatenation strategy.
In \emph{SOPA init}, + (\emph{resp.}, $\circ$) means that the background encoder is initialized with trained SOPA model (\emph{resp.}, ImageNet-pretrained ResNet-18). }
\label{table:component_table}
\centering
\begin{tabular}{c| c c c c c | c c } 
 \hline
    & dynamic filter & multi-scale & SOPA init & fix encoder & feature mimicking & F$_1$ $\uparrow$ &  bAcc$\uparrow$ \\  
 \hline
 1 & + & + & $\circ$  &   &   & 0.692 & 0.777 \\ 
 2 & + & + & $\circ$  & + &   & 0.713 & 0.787 \\ 
 3 & + & + & + &   &   & 0.694 & 0.778 \\ 
 4 & + & + & + & + &   & 0.737 & 0.810 \\ 
  \hline
 5 & + &   & + & + &   & 0.724 & 0.796 \\ 
 6 & $\dagger$ & + & + & + &   & 0.726 & 0.804 \\ 
 7 & $\dagger$ &   & + & + &   & 0.717 & 0.795 \\ 
 \hline
 8 & + & + & $\circ$  & + & + & 0.743 & 0.812 \\ 
 9 & + & + & + & + & + & 0.776 & 0.840 \\ 
 \hline
\end{tabular}
\end{table*}

\subsection{Dataset and Evaluation}
To satisfy our requirement, \emph{i.e.}, the rationality labels of composite images obtained by pasting foregrounds with different scales at different locations on the backgrounds, we evaluate our model on the recently released Object Placement Assessment (OPA) dataset \cite{liu2021opa}.
OPA dataset contains 73,470 composite images with rationality labels, which are split into 62,074 training images and 11,396 test images, where the foregrounds/backgrounds have no overlap between training set and test set. 
For each composite image, its composite foreground, foreground mask, and original background image are also provided.

To fit our model, we transform the image-level annotations in OPA to pixel-wise annotations. 
Specifically, given a background image and a foreground object of certain scale, we can collect the reasonable and unreasonable locations for this foreground based on the image-level annotations in OPA dataset. 
We refer to these locations as annotated pixels, in which the pixel-wise labels for the reasonable (\emph{resp.}, unreasonable) locations are set as $1$ (\emph{resp.}, $0$). More details can be found in Supplementary.
%and the other pixels as unlabeled pixels to get the target rationality score map for each item of our training data.
Considering that the foregrounds/backgrounds have no overlap between training and test set, we transform the training set and test set in OPA dataset separately. 
Furthermore, a pair of scaled foreground and background in the training (\emph{resp.}, test) set is dubbed as a training (\emph{resp.}, test) pair. 
Finally, we have 14,587 training pairs and 2,568 test pairs. 
Note that the pixel-wise annotations are very sparse and each pair only has 4 annotated pixels on average. 
We utilize these annotated pixels for training and testing. 

For evaluation metrics, following \cite{liu2021opa}, we adopt F$_1$ score and balanced accuracy (bAcc) based on the annotated pixels.

\subsection{Implementation Details}\label{sec:implementation}
We train our model for 20 epochs using Adam optimizer with a learning rate of 0.0005, and reduce the learning rate by a factor of 2 every 2 epochs. 
We ensure that the hyper-parameters and random seed do not change across experiments for consistency.  For the model input, background and foreground images are resized to $H\times W=256 \times 256$, so the size of rationality score map is also $256 \times 256$. We use the feature dimension $d=64$ and $\hat{d}=512$. We empirically set the dynamic kernel size $k=3$ and set the trade-off parameter $\lambda$ in Eqn. (\ref{eqn:total_loss}) as 16 via cross-validation.
Our network is implemented using Pytorch 1.9.0 and trained on Ubuntu 18.04.1 LTS, with 64GB memory, Intel(R) Xeon(R) CPU E5-2678 v3, and one GeForce GTX 1080 Ti GPU. 

\begin{figure*}[t]
\centering
\includegraphics[width=0.93\textwidth]{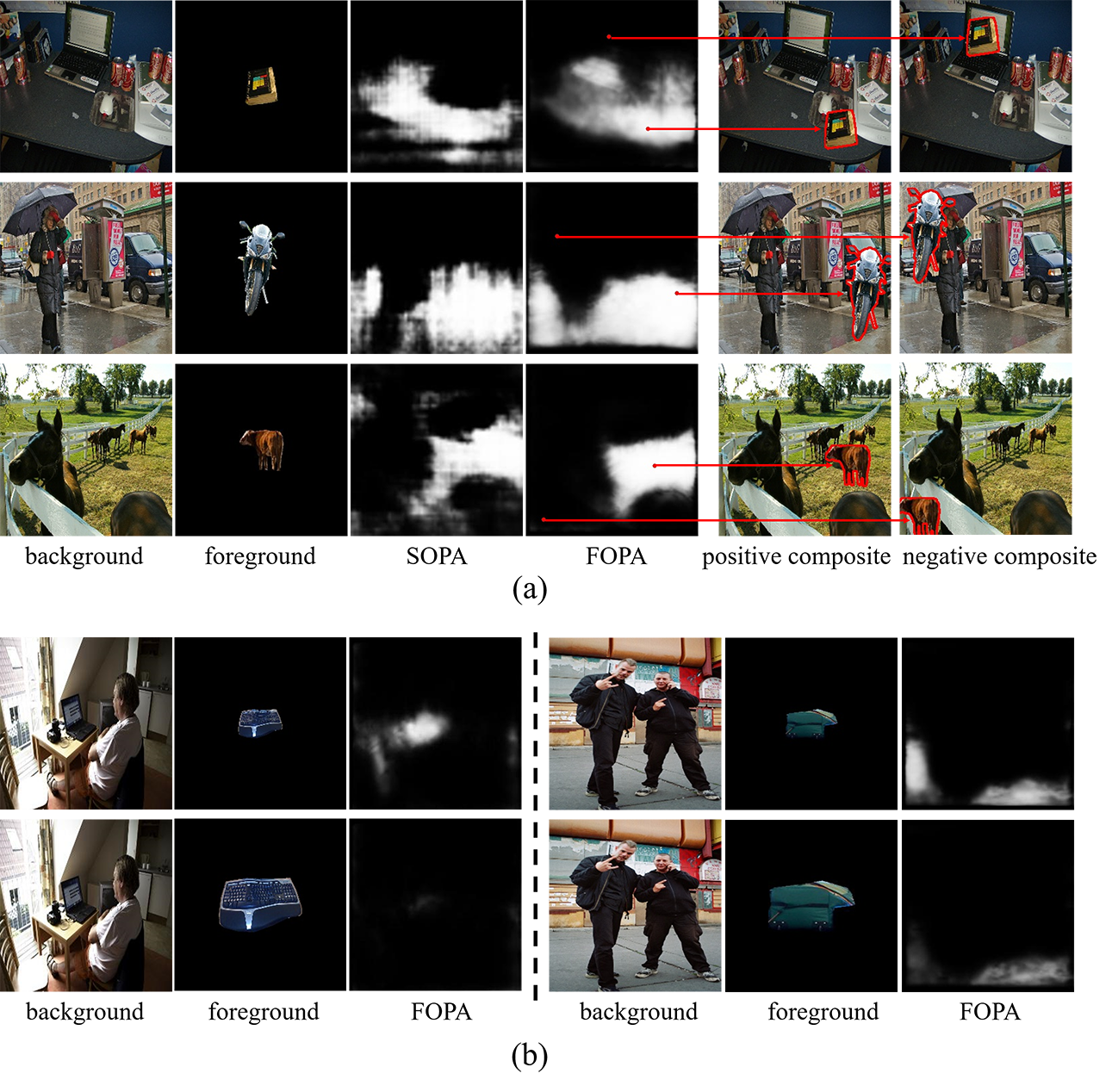} % Reduce the figure size so that it is slightly narrower than the column.

\caption{In (a), we compare the rationality score maps predicted by SOPA model and our FOPA model. We also display the positive (\emph{resp.}, negative) composite image with the highest (\emph{resp.}, lowest) rationality score. In (b), we show four triplets of background, foreground, and rationality score map predicted by our FOPA model to investigate the impact of foreground scale.}
\label{fig:Qualitative_Analyses}
\end{figure*}

\subsection{Ablation Study}
\label{section:Ablation_Studies}
In this section, we investigate the effectiveness of each design in our method and report the results in Table \ref{table:component_table}. 
We first study the initialization strategy and training strategy of background encoder $E_b$. 
For the initialization of $E_b$, we try two strategies: trained SOPA model and ImageNet-pretrained ResNet-18 (marked with $\circ$). 
We also try fixing or updating $E_b$ during training. 
As shown in Table \ref{table:component_table}, it is better to fix $E_b$ (row 1 \emph{v.s.} row 2, row 3 \emph{v.s.} row 4), which is helpful to prevent the overfitting issue. 
The comparison between row 2 and row 4 demonstrates that the background prior knowledge encapsulated in the trained SOPA model is useful for object placement assessment.

Apart from the dynamic filter, which is capable of integrating foreground feature and pixel-wise background feature, there are also some alternatives to achieve the same goal.
Here, we study another intuitive way to integrate foreground feature and pixel-wise background feature, \emph{i.e.}, concatenation (marked with $\dagger$). 
In detail, we spatially replicate the foreground feature vector and concatenate it with the background feature map, which is followed by a $1\times 1$ conv to produce the output feature map. We can also extend concatenation to multi-scale concatenation, similar to multi-scale dynamic filter.  
For both dynamic filter and concatenation, we implement multi-scale (2 scales) and single-scale versions. 
The results in row 4-7 show that dynamic filter is more effective than concatenation and multi-scale implementation can further boost the performance. 

Finally, we validate the effectiveness of feature mimicking. 
In row 1-7, we simply discard the feature mimicking loss by setting $\lambda$ in Eqn. (\ref{eqn:total_loss}) as 0. 
By comparing row 2 (\emph{resp.}, 4) and row 8 (\emph{resp.}, 9), we observe that feature mimicking can significantly improve the performance. 
The improvement based on SOPA initialization (row 9 \emph{v.s.} row 4) is larger, probably because it is easier to mimic the composite image features from SOPA model when using the trained SOPA model as the background encoder.

\begin{table*}[t]

\caption{Comparison between SOPA model and our proposed FOPA model. Time-s (\emph{resp.}, Memory) means the time (\emph{resp.}, required memory) of a single pass through the model. Time-a (\emph{resp.}, FLOPs) means the time (\emph{resp.}, FLOPs) required to predict a $256\times 256$ rationality score map.}

\label{tab:comp_SOPA_FOPA}
\centering
\begin{tabular}{c | c  c | c c c c} 
 \hline
 Method &  F$_1$ $\uparrow$ & bAcc$\uparrow$ & Time(s)-s$\downarrow$ & Time (s)-a$\downarrow$ & Memory (MB)$\downarrow$ & FLOPs (G)$\downarrow$ \\
 \hline
 SOPA & 0.780 & 0.842 & 0.0047 & 310.56 & 33.50 & 159,252.48 \\
 \hline
 FOPA & 0.776 & 0.840  & 0.0164 & 0.0164 & 279.90 & 31.94\\
 \hline
\end{tabular}
\end{table*}

\subsection{Comparison between FOPA and SOPA} \label{sec:comp_SOPA_FOPA}
To the best of our knowledge, SOPA \cite{liu2021opa} is the only existing discriminative approach for object placement, because the recently released OPA dataset \cite{liu2021opa} provides the possibility for researching on object placement assessment. The goal of this work is to accelerate SOPA while maintaining the effectiveness. 

\subsubsection{Effectiveness Comparison}

We first report F$_1$ and bAcc of SOPA and FOPA in
Table \ref{tab:comp_SOPA_FOPA}, from which we find that our FOPA model achieves comparable results with SOPA.

We visualize the predicted rationality score map of SOPA model and our FOPA model in the test set. 
For SOPA, we pass $256\times256$ composite images through the model one by one and concatenate $256\times256$ rationality scores as the entire rationality score map.
From Figure \ref{fig:Qualitative_Analyses}(a), we can see that both SOPA model and FOPA model can roughly predict the distribution of reasonable object locations. 
Based on the score map generated by FOPA model, we create the composite image with the highest (\emph{resp.}, lowest) score, by pasting the foreground at the locations with the highest (\emph{resp.}, lowest) pixel-wise rationality score. 
It is obvious that the composite image with the highest (\emph{resp.}, lowest) score is reasonable (\emph{resp.}, unreasonable). More visualization results can be found in Supplementary.  

For our FOPA method, we also investigate the impact of foreground scale on the predicted rationality score map. 
In Figure \ref{fig:Qualitative_Analyses}(b), it can be seen that the rationality score map changes dramatically as the size of the foreground changes. Some bright regions (\emph{e.g.}, desk) disappear when the foreground (\emph{e.g.}, keyboard) gets larger because there is not enough room to place the foreground any more.  

\subsubsection{Efficiency Comparison}
 
First, we test the inference speed on a single NVIDIA GTX 1080 Ti GPU. 
Specifically, we run SOPA and FOPA over all 2,568 test pairs and calculate the average time of a single inference pass through the model (Time-s).
FOPA is slower than SOPA because  SOPA is a simple encoder while FOPA has more complex network structure with a foreground branch and a background branch. 
However, a single pass of SOPA can only predict the rationality score for one pixel while a single pass of FOPA can predict the rationality scores for all pixels. 
Therefore, we compare the inference time per test pair (Time-a), that is, the time required to predict a $256\times 256$ rationality score map. The time of FOPA remains the same while the time of SOPA should be multiplied with the score map size $256\times 256$. 
It can be seen that FOPA model runs $19000\times $ faster than SOPA model, which demonstrates the efficiency of our FOPA model. 

We also report the memory cost and FLOPs. Although SOPA has smaller memory cost for each run due to simpler model structure, we need to run SOPA for $256\times 256$ times per test pair. The FLOPs per test pair of SOPA is dramatically larger than that of FOPA, which again proves the efficiency of our method. Therefore, our FOPA consumes much less resource than SOPA when producing a rationality score map. 

\subsection{Comparison with Generative Methods}
Recall that the existing object placement learning methods can be divided into discriminative approach and generative approach (see Section~\ref{sec:intro} and \ref{subsection:Object_Placement_Learning}). 
Although our focus is discriminative approach, we also compare with generative approaches for object placement assessment, because both types of approaches can be used to create reasonable composite images. 
We compare the quality of composite images generated by different discriminative approaches (SOPA and FOPA) and generative approaches (TERSE \cite{SyntheticTripathi2019} and PlaceNet \cite{LearningObjPlaZhang2020}) via user study.
Due to the space limitation, the visualization results and the user study details are left to Supplementary.

\section{Conclusion}
In this paper, we have defined a new task named fast object placement assessment (FOPA) and designed a pioneering FOPA model equipped with a series of innovations (\emph{e.g.}, foreground dynamic filter, background prior transfer, composite feature mimicking). The performance of our FOPA model is on par with slow object placement assessment (SOPA) model, but FOPA model runs significantly faster to produce a rationality score map, which could greatly benefit composite image assessment and composite image generation.

{\small
\bibliographystyle{ieee_fullname}
\bibliography{1main.bbl}

\begin{thebibliography}{10}\itemsep=-1pt

\bibitem{chawla2021data}
Akshay Chawla, Hongxu Yin, Pavlo Molchanov, and Jose Alvarez.
\newblock Data-free knowledge distillation for object detection.
\newblock In {\em WACV}, 2021.

\bibitem{TowardRealisticChen2019}
Bor{-}Chun Chen and Andrew Kae.
\newblock Toward realistic image compositing with adversarial learning.
\newblock In {\em {CVPR}}, 2019.

\bibitem{chen2017learning}
Guobin Chen, Wongun Choi, Xiang Yu, Tony Han, and Manmohan Chandraker.
\newblock Learning efficient object detection models with knowledge
  distillation.
\newblock {\em NIPS}, 2017.

\bibitem{chen2020dynamic}
Yinpeng Chen, Xiyang Dai, Mengchen Liu, Dongdong Chen, Lu Yuan, and Zicheng
  Liu.
\newblock Dynamic convolution: Attention over convolution kernels.
\newblock In {\em CVPR}, 2020.

\bibitem{chen2021geosim}
Yun Chen, Frieda Rong, Shivam Duggal, Shenlong Wang, Xinchen Yan, Sivabalan
  Manivasagam, Shangjie Xue, Ersin Yumer, and Raquel Urtasun.
\newblock Geosim: Realistic video simulation via geometry-aware composition for
  self-driving.
\newblock In {\em CVPR}, 2021.

\bibitem{CongDoveNet2020}
Wenyan Cong, Jianfu Zhang, Li Niu, Liu Liu, Zhixin Ling, Weiyuan Li, and Liqing
  Zhang.
\newblock Dovenet: Deep image harmonization via domain verification.
\newblock In {\em CVPR}, 2020.

\bibitem{IHCISSAMCun2020}
Xiaodong Cun and Chi-Man Pun.
\newblock Improving the harmony of the composite image by spatial-separated
  attention module.
\newblock {\em IEEE Transactions on Image Processing}, 29:4759--4771, 2020.

\bibitem{ding2020learning}
Mingyu Ding, Yuqi Huo, Hongwei Yi, Zhe Wang, Jianping Shi, Zhiwu Lu, and Ping
  Luo.
\newblock Learning depth-guided convolutions for monocular 3d object detection.
\newblock In {\em CVPR}, 2020.

\bibitem{MVCCPDvornik2018}
Nikita Dvornik, Julien Mairal, and Cordelia Schmid.
\newblock Modeling visual context is key to augmenting object detection
  datasets.
\newblock In {\em {ECCV}}, 2018.

\bibitem{InstaBoostFang2019}
Haoshu Fang, Jianhua Sun, Runzhong Wang, Minghao Gou, Yonglu Li, and Cewu Lu.
\newblock {InstaBoost}: Boosting instance segmentation via probability map
  guided copy-pasting.
\newblock In {\em ICCV}, 2019.

\bibitem{STDODISGeorgakis2017}
Georgios Georgakis, Arsalan Mousavian, Alexander~C. Berg, and Jana Kosecka.
\newblock Synthesizing training data for object detection in indoor scenes.
\newblock In {\em RSS}, 2017.

\bibitem{SynthesizingGeorgakis2017}
Georgios Georgakis, Arsalan Mousavian, Alexander~C. Berg, and Jana Kosecka.
\newblock Synthesizing training data for object detection in indoor scenes.
\newblock In {\em RSS}, 2017.

\bibitem{gou2021knowledge}
Jianping Gou, Baosheng Yu, Stephen~J Maybank, and Dacheng Tao.
\newblock Knowledge distillation: A survey.
\newblock {\em International Journal of Computer Vision}, 129(6):1789--1819,
  2021.

\bibitem{guo2019depthwise}
Yunhui Guo, Yandong Li, Liqiang Wang, and Tajana Rosing.
\newblock Depthwise convolution is all you need for learning multiple visual
  domains.
\newblock In {\em AAAI}, 2019.

\bibitem{guo2021intrinsic}
Zonghui Guo, Haiyong Zheng, Yufeng Jiang, Zhaorui Gu, and Bing Zheng.
\newblock Intrinsic image harmonization.
\newblock In {\em CVPR}, 2021.

\bibitem{he2016deep}
Kaiming He, Xiangyu Zhang, Shaoqing Ren, and Jian Sun.
\newblock Deep residual learning for image recognition.
\newblock In {\em CVPR}, 2016.

\bibitem{heo2019knowledge}
Byeongho Heo, Minsik Lee, Sangdoo Yun, and Jin~Young Choi.
\newblock Knowledge transfer via distillation of activation boundaries formed
  by hidden neurons.
\newblock In {\em Proceedings of the AAAI Conference on Artificial
  Intelligence}, volume~33, pages 3779--3787, 2019.

\bibitem{hinton2015distilling}
Geoffrey Hinton, Oriol Vinyals, and Jeff Dean.
\newblock Distilling the knowledge in a neural network.
\newblock {\em arXiv preprint arXiv:1503.02531}, 2015.

\bibitem{hong2021shadow}
Yan Hong, Li Niu, and Jianfu Zhang.
\newblock Shadow generation for composite image in real-world scenes.
\newblock In {\em {AAAI}}, 2022.

\bibitem{jia2016dynamic}
Xu Jia, Bert De~Brabandere, Tinne Tuytelaars, and Luc~V Gool.
\newblock Dynamic filter networks.
\newblock {\em NeurIPS}, 2016.

\bibitem{kim2018paraphrasing}
Jangho Kim, SeongUk Park, and Nojun Kwak.
\newblock Paraphrasing complex network: Network compression via factor
  transfer.
\newblock {\em arXiv preprint arXiv:1802.04977}, 2018.

\bibitem{ContextawareLee2018}
Donghoon Lee, Sifei Liu, Jinwei Gu, Ming{-}Yu Liu, Ming{-}Hsuan Yang, and Jan
  Kautz.
\newblock Context-aware synthesis and placement of object instances.
\newblock In {\em NeurIPS}, 2018.

\bibitem{Li_2017_CVPR}
Quanquan Li, Shengying Jin, and Junjie Yan.
\newblock Mimicking very efficient network for object detection.
\newblock In {\em CVPR}, July 2017.

\bibitem{Ling2021RegionawareAI}
Jun Ling, Han Xue, Li Song, Rong Xie, and Xiao Gu.
\newblock Region-aware adaptive instance normalization for image harmonization.
\newblock In {\em {CVPR}}, 2021.

\bibitem{ARShadowGANLiu2020}
Daquan Liu, Chengjiang Long, Hongpan Zhang, Hanning Yu, Xinzhi Dong, and
  Chunxia Xiao.
\newblock {ARShadowGAN}: Shadow generative adversarial network for augmented
  reality in single light scenes.
\newblock In {\em {CVPR}}, 2020.

\bibitem{liu2021opa}
Liu Liu, Bo Zhang, Jiangtong Li, Li Niu, Qingyang Liu, and Liqing Zhang.
\newblock {OPA}: Object placement assessment dataset.
\newblock {\em arXiv preprint arXiv:2107.01889}, 2021.

\bibitem{liu2019structured}
Yifan Liu, Ke Chen, Chris Liu, Zengchang Qin, Zhenbo Luo, and Jingdong Wang.
\newblock Structured knowledge distillation for semantic segmentation.
\newblock In {\em CVPR}, 2019.

\bibitem{niu2021making}
Li Niu, Wenyan Cong, Liu Liu, Yan Hong, Bo Zhang, Jing Liang, and Liqing Zhang.
\newblock Making images real again: A comprehensive survey on deep image
  composition.
\newblock {\em arXiv preprint arXiv:2106.14490}, 2021.

\bibitem{pang2020hierarchical}
Youwei Pang, Lihe Zhang, Xiaoqi Zhao, and Huchuan Lu.
\newblock Hierarchical dynamic filtering network for rgb-d salient object
  detection.
\newblock In {\em ECCV}, 2020.

\bibitem{passalis2020heterogeneous}
Nikolaos Passalis, Maria Tzelepi, and Anastasios Tefas.
\newblock Heterogeneous knowledge distillation using information flow modeling.
\newblock In {\em CVPR}, 2020.

\bibitem{LSCPRemez2018}
Tal Remez, Jonathan Huang, and Matthew Brown.
\newblock Learning to segment via cut-and-paste.
\newblock In {\em {ECCV}}, 2018.

\bibitem{RomeroBKCGB14}
Adriana Romero, Nicolas Ballas, Samira~Ebrahimi Kahou, Antoine Chassang, Carlo
  Gatta, and Yoshua Bengio.
\newblock Fitnets: Hints for thin deep nets.
\newblock In Yoshua Bengio and Yann LeCun, editors, {\em ICLR}, 2015.

\bibitem{ronneberger2015u}
Olaf Ronneberger, Philipp Fischer, and Thomas Brox.
\newblock U-net: Convolutional networks for biomedical image segmentation.
\newblock In {\em MICCAI}, 2015.

\bibitem{Sharma_2018_CVPR}
Vivek Sharma, Ali Diba, Davy Neven, Michael~S. Brown, Luc Van~Gool, and Rainer
  Stiefelhagen.
\newblock Classification-driven dynamic image enhancement.
\newblock In {\em CVPR}, 2018.

\bibitem{shu2021channel}
Changyong Shu, Yifan Liu, Jianfei Gao, Zheng Yan, and Chunhua Shen.
\newblock Channel-wise knowledge distillation for dense prediction.
\newblock In {\em ICCV}, 2021.

\bibitem{sofiiuk2021foreground}
Konstantin Sofiiuk, Polina Popenova, and Anton Konushin.
\newblock Foreground-aware semantic representations for image harmonization.
\newblock In {\em WACV}, 2021.

\bibitem{su2019pixel}
Hang Su, Varun Jampani, Deqing Sun, Orazio Gallo, Erik Learned-Miller, and Jan
  Kautz.
\newblock Pixel-adaptive convolutional neural networks.
\newblock In {\em CVPR}, 2019.

\bibitem{tabernik2020spatially}
Domen Tabernik, Matej Kristan, and Ale{\v{s}} Leonardis.
\newblock Spatially-adaptive filter units for compact and efficient deep neural
  networks.
\newblock {\em International Journal of Computer Vision}, 128(8):2049--2067,
  2020.

\bibitem{WhereandWhoTan2018}
Fuwen Tan, Crispin Bernier, Benjamin Cohen, Vicente Ordonez, and Connelly
  Barnes.
\newblock Where and who? {A}utomatic semantic-aware person composition.
\newblock In {\em WACV}, 2018.

\bibitem{tang2020learning}
Jie Tang, Fei-Peng Tian, Wei Feng, Jian Li, and Ping Tan.
\newblock Learning guided convolutional network for depth completion.
\newblock {\em TIP}, 30:1116--1129, 2020.

\bibitem{tian2020conditional}
Zhi Tian, Chunhua Shen, and Hao Chen.
\newblock Conditional convolutions for instance segmentation.
\newblock In {\em ECCV}, 2020.

\bibitem{SyntheticTripathi2019}
Shashank Tripathi, Siddhartha Chandra, Amit Agrawal, Ambrish Tyagi, James~M.
  Rehg, and Visesh Chari.
\newblock Learning to generate synthetic data via compositing.
\newblock In {\em CVPR}, 2019.

\bibitem{TsaiDIHarmonization2017}
Yi{-}Hsuan Tsai, Xiaohui Shen, Zhe Lin, Kalyan Sunkavalli, Xin Lu, and
  Ming{-}Hsuan Yang.
\newblock Deep image harmonization.
\newblock In {\em CVPR}, 2017.

\bibitem{wang2019carafe}
Jiaqi Wang, Kai Chen, Rui Xu, Ziwei Liu, Chen~Change Loy, and Dahua Lin.
\newblock Carafe: Content-aware reassembly of features.
\newblock In {\em ICCV}, 2019.

\bibitem{wang2021distilling}
Xiaochuan Wang, Aiguo Chen, Liang Zhang, Yi Gu, Mang Xu, and Haoyuan Yan.
\newblock Distilling the knowledge of multiscale densely connected deep
  networks in mechanical intelligent diagnosis.
\newblock {\em WCMC}, 2021, 2021.

\bibitem{wang2020exclusivity}
Xiaobo Wang, Tianyu Fu, Shengcai Liao, Shuo Wang, Zhen Lei, and Tao Mei.
\newblock Exclusivity-consistency regularized knowledge distillation for face
  recognition.
\newblock In {\em ECCV}, 2020.

\bibitem{wang2020advances}
Xinlong Wang, Rufeng Zhang, Tao Kong, Lei Li, and Chunhua Shen.
\newblock Solov2: Dynamic and fast instance segmentation.
\newblock In {\em NeurIPS}, 2020.

\bibitem{MISCWeng2020}
Shuchen Weng, Wenbo Li, Dawei Li, Hongxia Jin, and Boxin Shi.
\newblock {MISC:} {M}ulti-condition injection and spatially-adaptive
  compositing for conditional person image synthesis.
\newblock In {\em {CVPR}}, 2020.

\bibitem{yang2019condconv}
Brandon Yang, Gabriel Bender, Quoc~V Le, and Jiquan Ngiam.
\newblock Condconv: Conditionally parameterized convolutions for efficient
  inference.
\newblock {\em NeurIPS}, 2019.

\bibitem{yim2017gift}
Junho Yim, Donggyu Joo, Jihoon Bae, and Junmo Kim.
\newblock A gift from knowledge distillation: Fast optimization, network
  minimization and transfer learning.
\newblock In {\em CVPR}, 2017.

\bibitem{zhang2019fast}
Feng Zhang, Xiatian Zhu, and Mao Ye.
\newblock Fast human pose estimation.
\newblock In {\em CVPR}, 2019.

\bibitem{LearningObjPlaZhang2020}
Lingzhi Zhang, Tarmily Wen, Jie Min, Jiancong Wang, David Han, and Jianbo Shi.
\newblock Learning object placement by inpainting for compositional data
  augmentation.
\newblock In {\em ECCV}, 2020.

\bibitem{zhang2017scale}
Rui Zhang, Sheng Tang, Yongdong Zhang, Jintao Li, and Shuicheng Yan.
\newblock Scale-adaptive convolutions for scene parsing.
\newblock In {\em ICCV}, 2017.

\bibitem{WhatWhereZhang2020}
Song{-}Hai Zhang, Zhengping Zhou, Bin Liu, Xi Dong, and Peter Hall.
\newblock What and where: {A} context-based recommendation system for object
  insertion.
\newblock {\em Computational Visual Media}, 6(1):79--93, 2020.

\bibitem{zhou2021decoupled}
Jingkai Zhou, Varun Jampani, Zhixiong Pi, Qiong Liu, and Ming-Hsuan Yang.
\newblock Decoupled dynamic filter networks.
\newblock In {\em CVPR}, 2021.

\end{thebibliography}


\begin{thebibliography}{1}\itemsep=-1pt

\bibitem{liu2021opa}
Liu Liu, Bo Zhang, Jiangtong Li, Li Niu, Qingyang Liu, and Liqing Zhang.
\newblock {OPA}: Object placement assessment dataset.
\newblock {\em arXiv preprint arXiv:2107.01889}, 2021.

\bibitem{SyntheticTripathi2019}
Shashank Tripathi, Siddhartha Chandra, Amit Agrawal, Ambrish Tyagi, James~M.
  Rehg, and Visesh Chari.
\newblock Learning to generate synthetic data via compositing.
\newblock In {\em CVPR}, 2019.

\bibitem{LearningObjPlaZhang2020}
Lingzhi Zhang, Tarmily Wen, Jie Min, Jiancong Wang, David Han, and Jianbo Shi.
\newblock Learning object placement by inpainting for compositional data
  augmentation.
\newblock In {\em ECCV}, 2020.

\end{thebibliography}
}

\end{document}

%%%%%%%%% TITLE
%%%%%%%%% TITLE
\title{Supplementary for Fast Object Placement Assessment}

\author{$\textnormal{Li Niu}^{*}$, $\textnormal{Qingyang Liu}^{\dag}$, $\textnormal{Zhenchen Liu}^{*}$, $\textnormal{Jiangtong Li}^{*}$\\
$^*$ Shanghai Jiao Tong University\,\,
$\dag$ Beijing Institute of Technology \\
}

\maketitle
In this document, we provide additional materials to support our main paper. In Section \ref{sec:pixel_annotation}, we explain the conversion from image-level annotation to pixel-wise annotation in detail. 
In Section \ref{sec:Foreground_Scale}, we explore an alternative approach to encode the foreground scale information, which can accelerate processing multiple foreground scales. In Section \ref{sec:More_Visualization_Results}, we show more visualization results of SOPA and FOPA heatmaps. In Section \ref{sec:compare_with_generative},  we compare with generative baselines via qualitative results and user study.

\section{From Image-level Annotation to Pixel-wise Annotation}\label{sec:pixel_annotation}

For slow object placement assessment (OPA) task, the input is a composite image and its composite foreground mask, while the output is image-level binary rationality score. The recently released Object Placement Assessment (OPA) dataset \cite{liu2021opa} contains 73,470 composite images with rationality labels, which can be directly used for slow OPA task. For fast object placement assessment (OPA) task, the input is a foreground with certain scale and its foreground mask as well as a background image, while the output is a rationality score map with the predicted rationality score for each location. 

\begin{figure*}[h]
\centering
\includegraphics[width=0.9\textwidth]{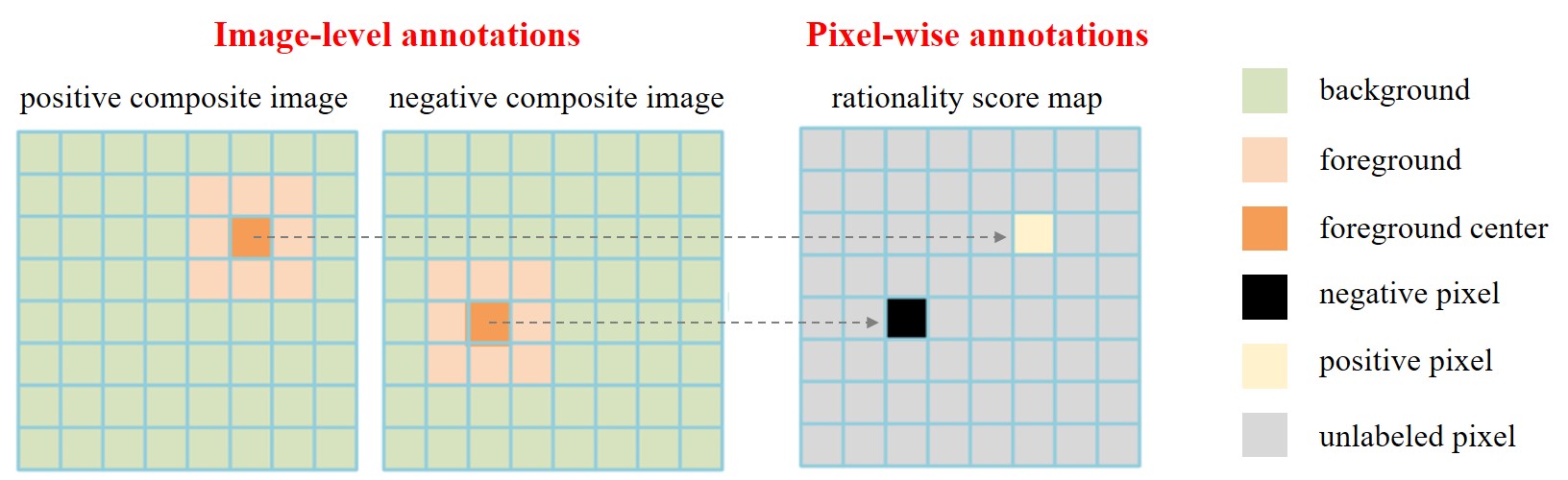} % Reduce the figure size so that it is slightly narrower than the column.

\caption{The illustration of image-level annotations for composite images, and pixel-wise annotations for a pair of scaled foreground and background. The image-level label of the composite image obtained by placing the scaled foreground at certain location on the background is equal to the pixel-wise label at this location in the rationality score map. }

\label{fig:pixel_annotation}
\end{figure*}

Therefore, we need to convert the image-level annotations in OPA dataset to pixel-wise annotations for fast OPA task, as shown in Figure~\ref{fig:pixel_annotation}. Specifically, given a pair of scaled foreground and background, we can collect the reasonable and unreasonable locations for this foreground based on the image-level annotations in OPA dataset, corresponding to positive pixels and negative pixels respectively. The other locations without annotations correspond to unlabeled pixels. The positive pixels, negative pixels, and unlabeled pixels form the entire rationality score map. 

\section{Encoding Foreground Scale Information}
\label{sec:Foreground_Scale}
In our method, we include the foreground scale information by directly scaling the foreground image. 
Specifically, we place the scaled foreground object $\mathbf{O}$ and mask $\mathbf{M}$ at the centre of a $H\times W$ black image, resulting in the input foreground $\mathbf{O}'\in\mathcal{R}^{H\times W}$ and $\mathbf{M}'\in\mathcal{R}^{H\times W}$ . However, one drawback of this strategy is that we need to pass through the foreground encoder $N_s$ times for $N_s$ foreground scales. 

In this section, we explore an alternative approach to encode the foreground scale information, which can accelerate processing different foreground scales. 
Specifically, we resize $\mathbf{O}$ and $\mathbf{M}$ to fill up $H\times W$ input so that the foreground object fully occupies the input image, in which more details could be preserved but the scale information is lost. Then, we employ an one-hot vector to present the quantized foreground scale, which is concatenated with the output feature of foreground encoder.  
Precisely, we discretize foreground scale into $b$ bins and use a $b$-dim one-hot vector to indicate the corresponding bin. 
We vary $b$ in the range of [8, 16, 32] and report the results in Table \ref{table:foreground_scale}. For one-hot encoding, larger $b$ generally performs better, but the results are still worse than scaling the foreground directly. Therefore, we can choose the suitable strategy to encode foreground scale information according to the requirement of efficiency and effectiveness. 

\begin{table}[t]
\caption{We compare two approaches to encode foreground scale information: ``Ours" and ``One-hot Scale Encoding" with different numbers of bins. }
\label{table:foreground_scale}
\centering
\begin{tabular}{c | c | r | r | r } 
 \hline
 \multirow{2}*{ } &
 \multirow{2}*{Ours} &
 \multicolumn{3}{|c}{One-hot Scale Encoding} \\ \cline{3-5}
 & & 8\quad\quad & 16\quad\quad & 32\quad\quad \\
 \hline
 F$_1$ & 0.776 & 0.732 & 0.744 & 0.735  \\ 
 bAcc & 0.840 & 0.803 & 0.812 & 0.808 \\ 
 \hline
\end{tabular}
\end{table}

\begin{figure*}[t]
\centering
\includegraphics[width=0.9\textwidth]{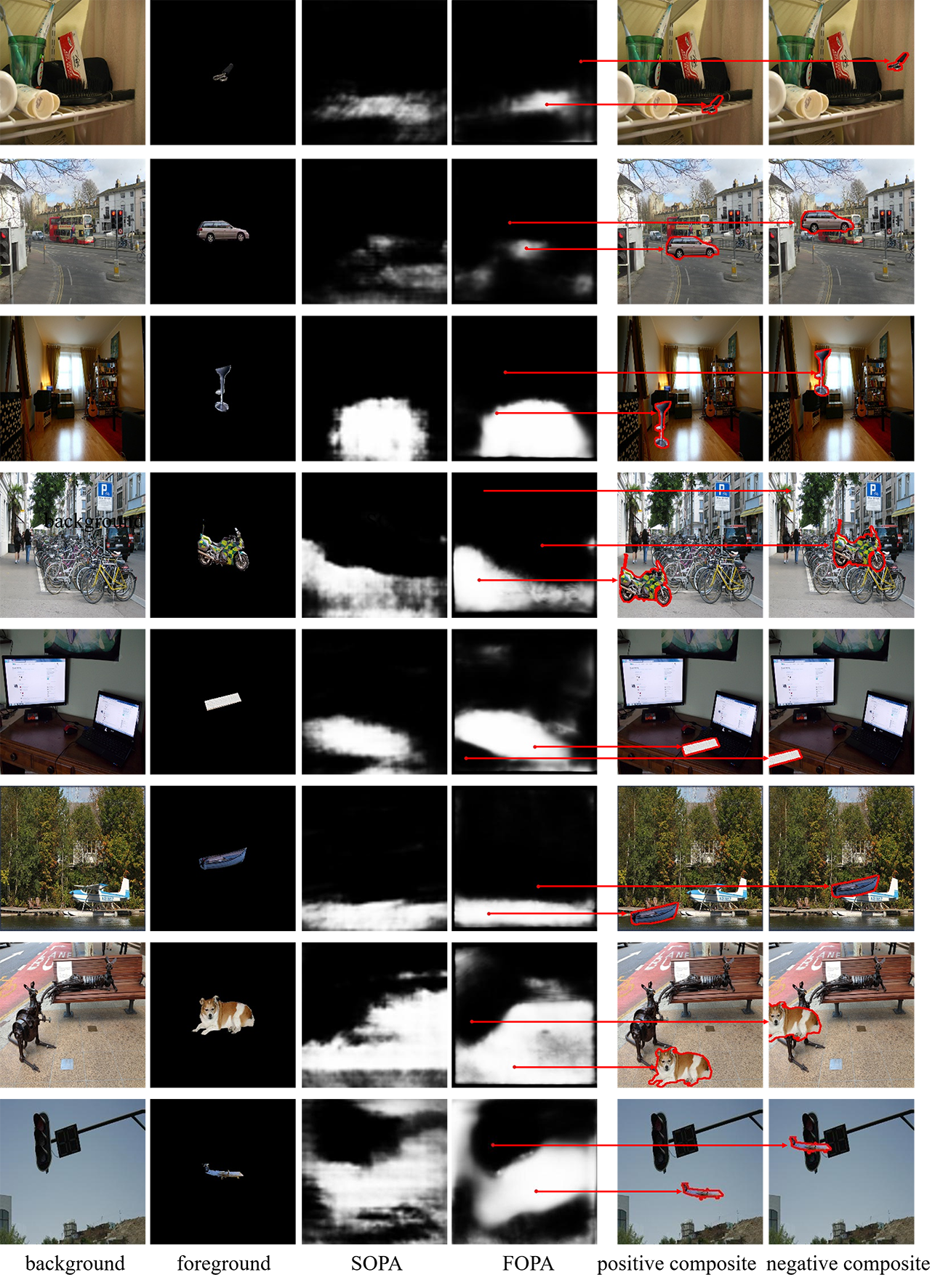} % Reduce the figure size so that it is slightly narrower than the column.

\caption{We compare the rationality score maps predicted by SOPA model and our FOPA model. We also display the positive (\emph{resp.}, negative) composite image with the highest (\emph{resp.}, lowest) rationality score.}

\label{fig:Visualization_heatmap}
\end{figure*}

\begin{figure*}[h]
\centering
\includegraphics[width=0.9\textwidth]{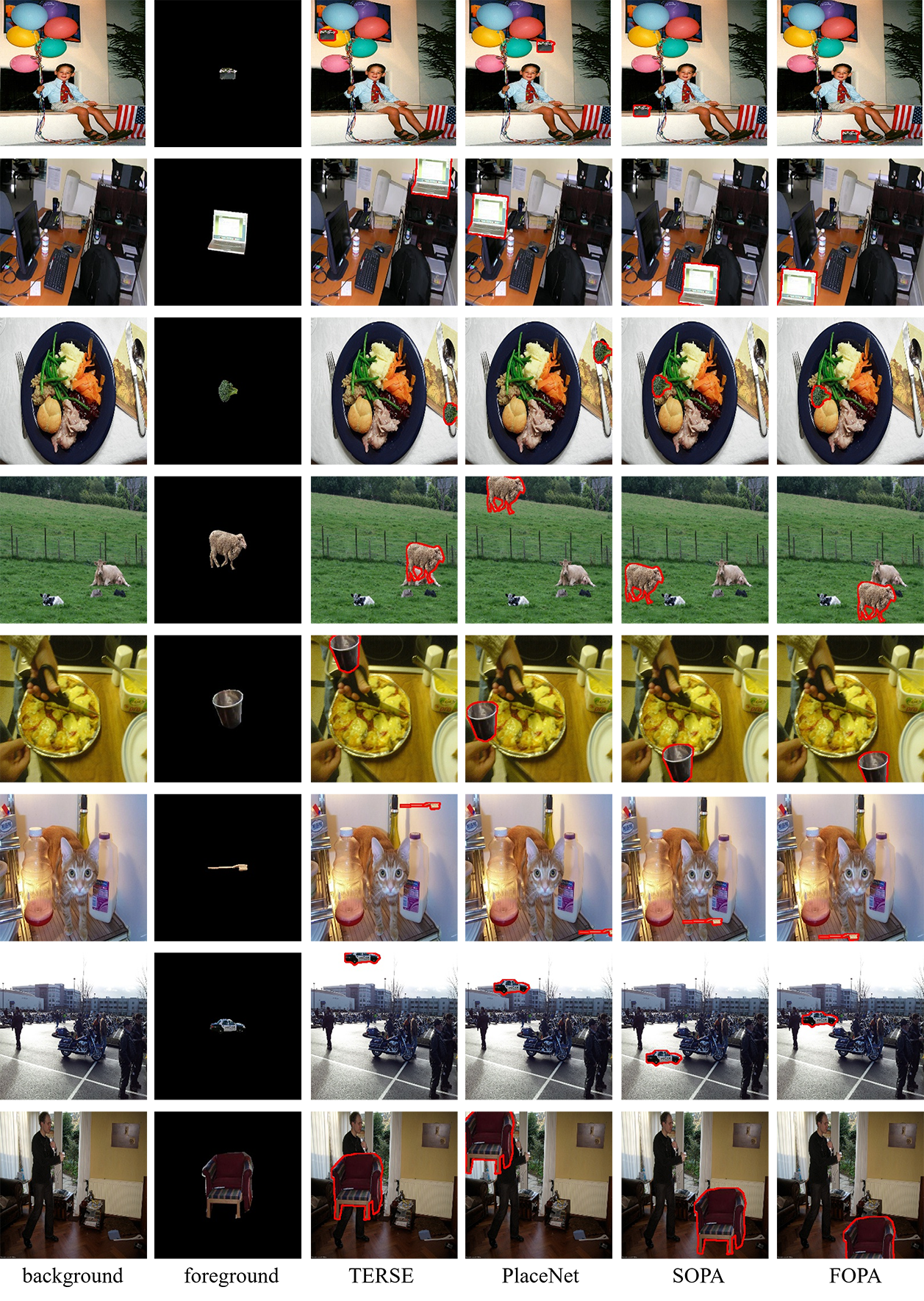} % Reduce the figure size so that it is slightly narrower than the column.

\caption{We compare the positive composite images generated by SOPA, FOPA, TERSE, and PlaceNet.}

\label{fig:Visualization_four_methord}
\end{figure*}

\section{More Visualization Results of SOPA and FOPA Heatmaps}
\label{sec:More_Visualization_Results}

Similar to Section 4.4 of the main paper, we first visualize the predicted rationality score map of SOPA model and our FOPA model in the test set in Figure \ref{fig:Visualization_heatmap}. Specifically, SOPA model enumerates all foreground locations and predicts the rationality scores of obtained composite images, after which the rationality scores of all composite images constitute the complete rationality score map. In contrast, our FOPA model only needs to pass the foreground and background through the model once to yield a rationality score map. As shown in Figure \ref{fig:Visualization_heatmap}, both SOPA model and FOPA model can roughly predict the distribution of reasonable object locations. One observation is that our FOPA model can choose the semantically suitable area for the foreground object, for example, scissor on the shelf (row 1), keyboard on the table (row 5), boat on the lake (row 6), airplane in the sky (row 8). Another observation is that besides semantically suitable area, our FOPA model also takes depth and foreground scale into consideration when determining the reasonable location. For example, in row 2, our FOPA heatmap locates the distant road surface instead of covering all the road surfaces. Similar observations could be found in row 3 in Figure3(a) of the main paper, \emph{e.g.}, our FOPA heatmap locates the nearby grassland while ignoring the distant grassland.

Based on the score map generated by our FOPA model, we create the positive (\emph{resp.}, negative) composite image with the highest (\emph{resp.}, lowest) score, by pasting the foreground at the location with the highest (\emph{resp.}, lowest) pixel-wise rationality score. The positive and negative composite images are shown in Figure \ref{fig:Visualization_heatmap}. For positive composite images, the foreground is placed at reasonable location and the whole image looks realistic. For negative composite image, the foreground is floating in the air without supporting force, or has unreasonable occlusion, or appears at semantically unreasonable location, which makes the whole image apparently unrealistic. 
 
\section{Comparison with Generative Methods} \label{sec:compare_with_generative}
Although our focus is discriminative approach, we also compare with generative approaches for object placement assessment, because both types of approaches can be used to create reasonable composite images. 
For discriminative models SOPA and FOPA, we obtain the positive composite image according to the highest pixel-wise rationality score in the score map, as described in Section~\ref{sec:More_Visualization_Results}. 
For generative models TERSE \cite{SyntheticTripathi2019} and PlaceNet \cite{LearningObjPlaZhang2020}, we generate a reasonable placement and obtain the positive composite image. Note that FOPA and SOPA produce a rationality score map for a pair of scaled foreground and background. For fair comparison, when employing generative models TERSE and PlaceNet, we fix the foreground size and only predict a reasonable location for the foreground. 
We show the positive composite images generated by SOPA, FOPA, TERSE, and PlaceNet in Figure~\ref{fig:Visualization_four_methord}.  We can see that the SOPA model and our FOPA model can generate more realistic composite images than TERSE and PlaceNet, while our FOPA runs $19000\times$ faster than SOPA.

Furthermore, we compare the positive composite images generated by SOPA, FOPA, TERSE, and PlaceNet via user study. Specifically, we conduct user study on the 2,568 pairs of scaled foregrounds and backgrounds in the test set.  For each test pair, we construct a group of four composite images which contain the positive composite images generated by SOPA, FOPA, TERSE, and PlaceNet. We invite 20 human raters to select the most realistic image from each group. Then, we calculate the percentage that the positive composite image generated by each method is selected as the most realistic one. The percentages of SOPA, FOPA, TERSE, and PlaceNet are $30.8\%$, $30.2\%$,  $19.8\%$, and  $19.2\%$, respectively. 
Overall, the results indicate that the SOPA model and our FOPA model can generate more realistic images.

{\small
\bibliographystyle{ieee_fullname}
\bibliography{2supplementary.bbl}
}